# Progressive Knowledge-Guided Large Language Model Framework for Bearing Fault Diagnosis

Jinghan Wang[a], Gaoliang Peng[a]*, Yanjun Chen[b], Wei Zhang[b], Wentao Wu[a] and Tianchen Liu[a]

[a] Harbin Institute of Technology, China
[b] Eastern Institute of Technology, China

**Abstract**
Vibration-based bearing fault diagnosis requires resolving three interrelated measurement challenges, including the trade-off between global statistical feature efficiency and local transient signal fidelity, insufficient traceability of measurement features to underlying fault physics, and ineffective multi-source measurement information fusion across diagnostic scales. This paper presents a progressive physics-guided multi-scale vibration signal processing framework that addresses all three challenges within a unified diagnostic pipeline. An 81-dimensional measurement descriptor, derived from bearing kinematic theory and characteristic defect frequencies, establishes a physically traceable feature space enabling real-time fault screening at approximately 20 ms per sample. A fault-adaptive signal segmentation mechanism then directs analytical attention toward fault-relevant waveform regions guided by physics-based priors, without manual feature engineering. Structured fault mechanism knowledge is further encoded implicitly in model parameters during training, enabling autonomous multi-scale measurement fusion without external knowledge dependencies at inference. Validated on four public benchmark datasets under diverse operating conditions, the framework achieves 98.49% diagnostic accuracy with a 12.6-fold reduction in computational cost relative to signal-level baselines. Interpretability analysis confirms that diagnostic feature activations align with established bearing fault mechanics, supporting measurement traceability in safety-critical industrial systems.

## 1. Introduction

Rolling element bearing faults account for 40% to 90% of all rotating machinery failures [1]. In wind turbines alone, bearing faults constitute 40% of component failures and represent the most frequent failure mode [2]. The overarching objective of bearing fault diagnosis is to identify fault type and severity at the earliest detectable stage, given vibration signals acquired continuously from bearings operating under variable load and speed conditions. Meeting this objective underpins the transition from reactive to condition-based predictive maintenance, which substantially reduces unplanned downtime, extends component lifespan, and optimises maintenance scheduling [3], [4]. Industrial deployment, however, imposes constraints that extend well beyond classification accuracy. Diagnostic decisions must be reached within latencies compatible with online monitoring, must remain auditable against established bearing fault mechanics to satisfy the interpretability requirements of safety-critical operations, and must be executable without dependence on network-accessible external knowledge repositories that are unavailable in many field installations. Despite two decades of research spanning signal processing, machine learning, and deep learning paradigms, no existing framework simultaneously satisfies all three of these deployment constraints, and a fully satisfactory solution remains elusive.

The initial generation of diagnostic methods combined hand-crafted spectral features with shallow classifiers, drawing on signal processing techniques such as Fast Fourier Transform [5], wavelet decomposition [6], and variational mode decomposition [7], combined with machine learning classifiers including Support Vector Machines [8], [9] and Random Forests [10]. Although physically interpretable and computationally efficient, these methods require substantial domain expertise for feature construction and remain sensitive to operating-condition variability, constraining their scalability across diverse industrial settings.

To alleviate the feature engineering bottleneck, deep learning architectures were introduced for automatic representation learning. Convolutional Neural Networks extracted spatial patterns from time-frequency maps [11], [12], while Recurrent Neural Networks and Long Short-Term Memory variants captured temporal dependencies in vibration signals [13], [14], and Transformer-based models exploited multi-scale contextual relationships through self-attention [15], [16]. These architectures, however, introduce two structural limitations that are directly relevant to industrial deployment. Feature-based architectures discard transient fault signatures through signal compression, whereas end-to-end signal-level processing incurs computational demands that are prohibitive for real-time online monitoring [17]. More fundamentally, purely data-driven models encode no explicit fault mechanism knowledge, precluding the post-hoc auditability required for regulatory acceptance in safety-critical industrial systems [18], [19], [20], [21].

Recognizing this interpretability deficit, a separate line of research has incorporated fault mechanism knowledge graphs (KGs) [22], [23], [24], and domain-specific ontologies [25] to encode structured physical priors. While these representations enhance interpretability, they remain static under novel operating conditions, and their integration has typically been confined to post-hoc annotation rather than active guidance of feature extraction, limiting diagnostic performance under variable industrial conditions. Furthermore, the majority of KG-enhanced methods rely on runtime queries to external knowledge repositories, introducing inference-time latency and infrastructure dependencies that are incompatible with autonomous on-site deployment, where continuous monitoring must proceed without access to centralised knowledge services.

The emergence of Large Language Models (LLMs) has opened a qualitatively different avenue, offering cross-domain reasoning capabilities valuable for adaptive knowledge integration. Two broad strategies have been pursued. The first integrates LLMs with structured knowledge sources, where agent-based reasoning over KGs [26] and prefix-tuning with subgraph embedding [27] have been employed to enable adaptive knowledge utilization during inference. The second focuses on cross-dataset generalization through signal-to-language conversion, where learnable prompts combined with Low-Rank Adaptation [28] and signal textualization with parameter-efficient fine-tuning [29] have demonstrated improved transferability across operating conditions. BearLLM [30] pursues a complementary direction, incorporating prior knowledge through signal-level representation engineering via adaptive sampling and frequency-domain unification.

A review of the foregoing literature indicates that three fundamental and interrelated challenges remain unresolved in intelligent bearing fault diagnosis for industrial deployment. The first challenge is Scale Conflict. Accurate diagnosis requires both global statistical patterns and local fault-induced impulses. Feature representations are computationally efficient but lose local transient detail, whereas raw signal processing preserves temporal fidelity at a computational cost that is prohibitive for real-time deployment. No existing framework captures both within a single deployable architecture. Another challenge is the Knowledge Gap. Data-driven models, including LLM-based approaches, lack explicit encoding of bearing dynamics and fault mechanism priors. KG integration has remained superficial, functioning as post-hoc annotation rather than actively guiding feature extraction and attention allocation throughout the diagnostic process. The final challenge is Multi-scale Integration Insufficiency. Independently staged pipelines without inter-stage information flow prevent global feature-based diagnostic priors from actively constraining local patch-based signal analysis, limiting both diagnostic accuracy and interpretability under variable operating conditions.

Considered together, these challenges manifest as a deployment gap that no existing framework has closed. In industrial monitoring practice, a diagnostic system must simultaneously achieve latencies compatible with online operation, produce decisions that are traceable to known bearing fault mechanisms, and operate without runtime dependence on external knowledge infrastructure. Prior LLM-based methods do not satisfy this combination of requirements. Signal textualization approaches sacrifice the temporal fidelity that is essential for capturing transient fault impulses, while KG-integrated methods retain runtime retrieval dependencies that are impractical in field-deployed systems. The present work directly targets this deployment gap through a principled integration of physical domain knowledge into every stage of the diagnostic process.

To close this gap, a three-stage progressive physics-guided vibration signal processing framework is proposed for interpretable bearing fault diagnosis, advancing from global measurement feature analysis through fault-aware local signal segmentation to multi-modal measurement fusion, without introducing external knowledge dependencies at inference time. The main contributions of this work are as follows.

1) A knowledge-enhanced 81-dimensional multi-scale measurement descriptor is constructed from bearing kinematic theory, integrating time-domain statistical descriptors of rolling element-raceway contact events, spectral energy components reflecting structural resonance under fault-induced loading, and frequency-domain descriptors derived from Ball Pass Frequency Inner race, Ball Pass Frequency Outer race, and Ball Spin Frequency. Each descriptor maintains a one-to-one correspondence with its fault source periodicity, establishing direct traceability between measurement features and fault mechanisms.
2) A fault-adaptive signal segmentation mechanism is introduced in which analytical attention over raw vibration segments is dynamically modulated by the Stage 1 fault estimate, guided by the known temporal periodicity and waveform asymmetry of bearing fault impulse responses. This directs analytical focus toward mechanistically expected fault intervals without manual signal pre-processing.
3) A three-stage progressive measurement-to-diagnosis pipeline is designed in which the Stage 1 mechanistically grounded fault prior actively constrains subsequent signal analysis and fusion computations, achieving approximately 20 milliseconds per-sample screening latency and a 12.6-fold computational cost reduction relative to signal-level baselines, satisfying the efficiency constraints of online industrial monitoring.

Together, these contributions show that grounding a multi-scale vibration signal processing framework in bearing physics improves diagnostic accuracy, generalizes across operating conditions, and remains practical for real industrial deployment. The remainder of this paper is organized as follows. Section 2 presents the proposed framework. Section 3 describes the experimental setup and results. Section 4 discusses generalizability and practical implications. Section 5 concludes the paper.

## 2. Preliminary and Problem Formulation

This work addresses the bearing fault diagnosis problem within a supervised learning framework. Given a raw vibration signal $\boldsymbol{x(t)} \in \mathbb{R}^N$, where $N$ denotes the number of sampling points, the objective is to learn a diagnostic function:

$$f : \boldsymbol{x} \to y \,, \tag{1}$$

that maps the input signal to a fault category $y \in Y$. The label space $Y$ comprises either four classes $\{Normal, Inner\ Race, Outer\ Race, Ball\}$ or three classes $\{Normal, Inner\ Race, Outer\ Race\}$, depending on the dataset characteristics.

The diagnostic function $f$ must capture multi-scale fault signatures, from global spectral characteristics to localized transient anomalies, while maintaining consistency with bearing dynamics theory to ensure physically plausible diagnoses. Industrial deployment further requires diagnostic interpretability through interpretable decision pathways traceable to domain knowledge. Satisfying these requirements necessitates integrating heterogeneous information sources, including signal representations, localized patterns, and structured domain constraints, within a unified framework for robust cross-condition generalization.

The learning objective is formulated as minimizing the classification loss:

$$\mathcal{L}(f(\boldsymbol{x};\boldsymbol{\theta}), y), \tag{2}$$

where $\theta$ denotes the model parameters. Critically, this optimization is subject to interpretability constraints imposed by domain knowledge, ensuring that the learned diagnostic function $f$ aligns with established bearing fault mechanism principles. This knowledge-guided framework enables robust generalization across diverse operating conditions while maintaining diagnostic interpretability essential for industrial deployment.

## 3. Proposed Methodology

### 3.1 Overall framework

Effective bearing fault diagnosis requires balancing two competing objectives: global pattern recognition for efficient fault categorization and local detail preservation for accurate fault characterization. Conventional approaches typically prioritize one objective at the expense of the other, resulting in either limited discriminative power or excessive computational cost. To address this fundamental trade-off, this research proposes a three-stage progressive diagnosis framework that systematically integrates knowledge-guided feature engineering, patch-based signal modelling, and multi-modal fusion within a unified LLM architecture.

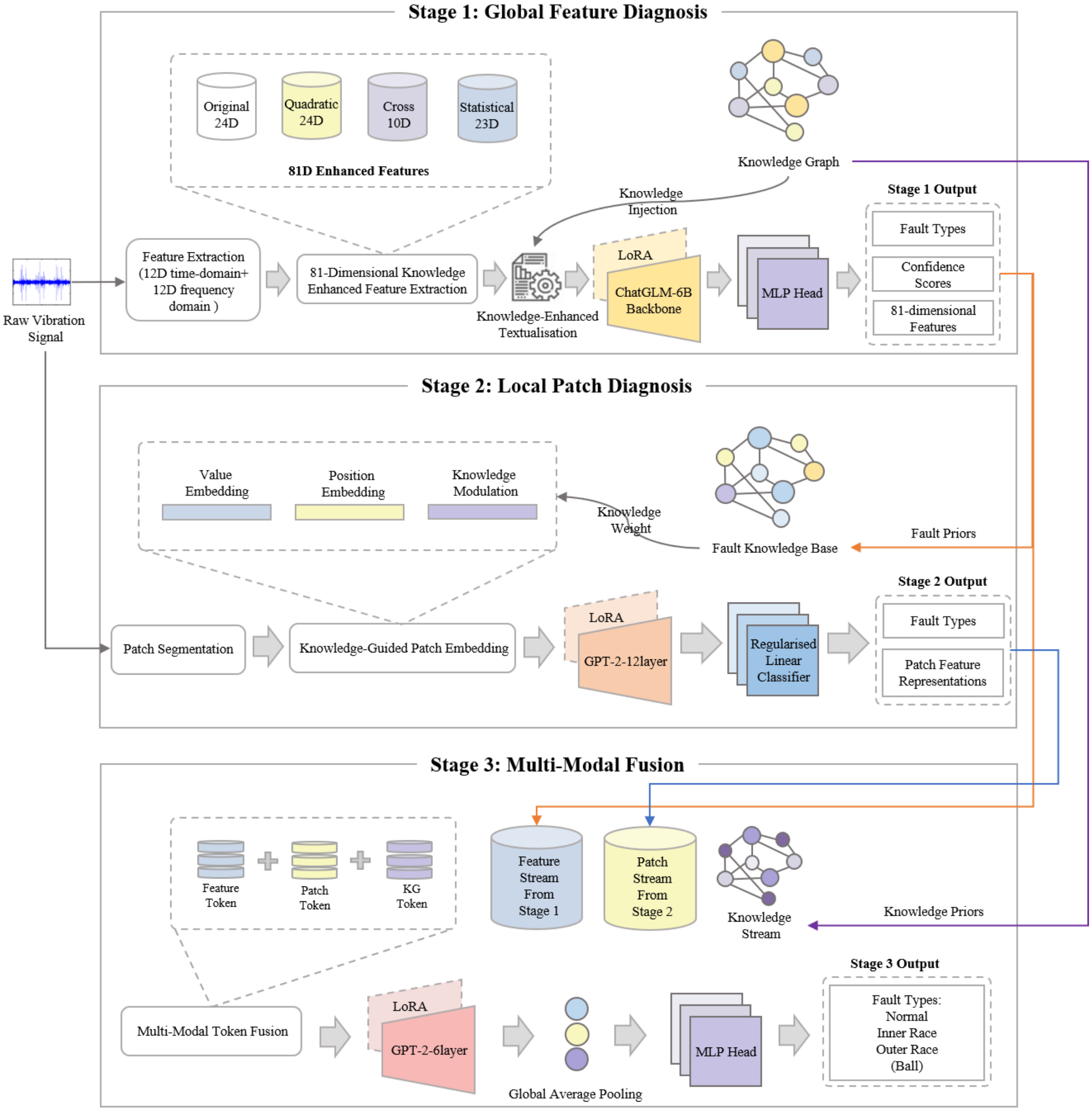

**Fig. 1.** Three-Stage LLM-Based Progressive Diagnosis Framework.

As illustrated in Fig. 1, the framework comprises three sequential yet complementary stages. Given a raw vibration signal $x(t)$, the diagnostic process proceeds as follows.

**Stage 1: Global Feature Diagnosis** extracts an 81-dimensional knowledge-enhanced feature vector $f_{global}$ from the input signal, encompassing time-domain, frequency-domain, and wavelet-domain statistics grounded in bearing dynamics theory. These features are converted into structured textual descriptions and processed by a fine-tuned ChatGLM-6B model with LoRA adaptation, yielding a feature-based fault classification $y_1$ and a fault prior $p_1$ that encapsulates probabilistic knowledge about the likely fault type to guide subsequent stages.

**Stage 2: Local Patch Diagnosis** segments the raw signal into overlapping patches $\{x_i\}_{i=1}^{M}$ that preserve temporal continuity. Patch embeddings are modulated by knowledge-guided attention weights $\alpha(p_1)$ derived from the Stage 1 prior $p_1$, reflecting fault-mechanism-specific characteristics. A 12-layer GPT-2 backbone with LoRA processes these modulated embeddings, producing a patch-based classification $y_2$ that captures local transient dynamics.

**Stage 3: Multi-Modal Integration** consolidates complementary information streams by constructing a heterogeneous token sequence $[t_{\{global\}}, t_{\{patch,1\}}, \ldots, t_{\{patch,M\}}, t_{\{KG\}}]$, where $t_{\{global\}}$ represents global features from Stage 1, $t_{\{patch,i\}} (i = 1, \ldots, M)$ denote local patch embeddings from Stage 2, and $t_{\{KG\}}$ incorporates external KG representations encoding bearing geometry and fault mechanisms. A lightweight 6-layer GPT-2 model processes this unified sequence through end-to-end optimization, yielding the final diagnostic prediction $y_{final}$.

## 3.2 Stage 1: global feature diagnosis

As shown in Fig. 2, Stage 1 implements knowledge-enhanced feature-level diagnosis by transforming raw vibration signals into semantic predictions. Raw acceleration data undergoes time-frequency feature extraction, producing 24 fundamental descriptors. These are expanded to 81 dimensions through four components: original 24 features, 24 quadratic terms for nonlinear patterns, 10 cross-feature interactions for synergistic relationships, and 23 higher-order statistics for global characteristics. Simultaneously, an external KG containing diagnostic templates for four fault types is queried via cosine similarity, yielding prior probability scores. The 81-dimensional features and knowledge scores are converted to natural language prompts, processed by LoRA-adapted ChatGLM-6B, and classified via an MLP head to produce fault predictions, confidence scores, and refined feature embeddings for Stage 2.

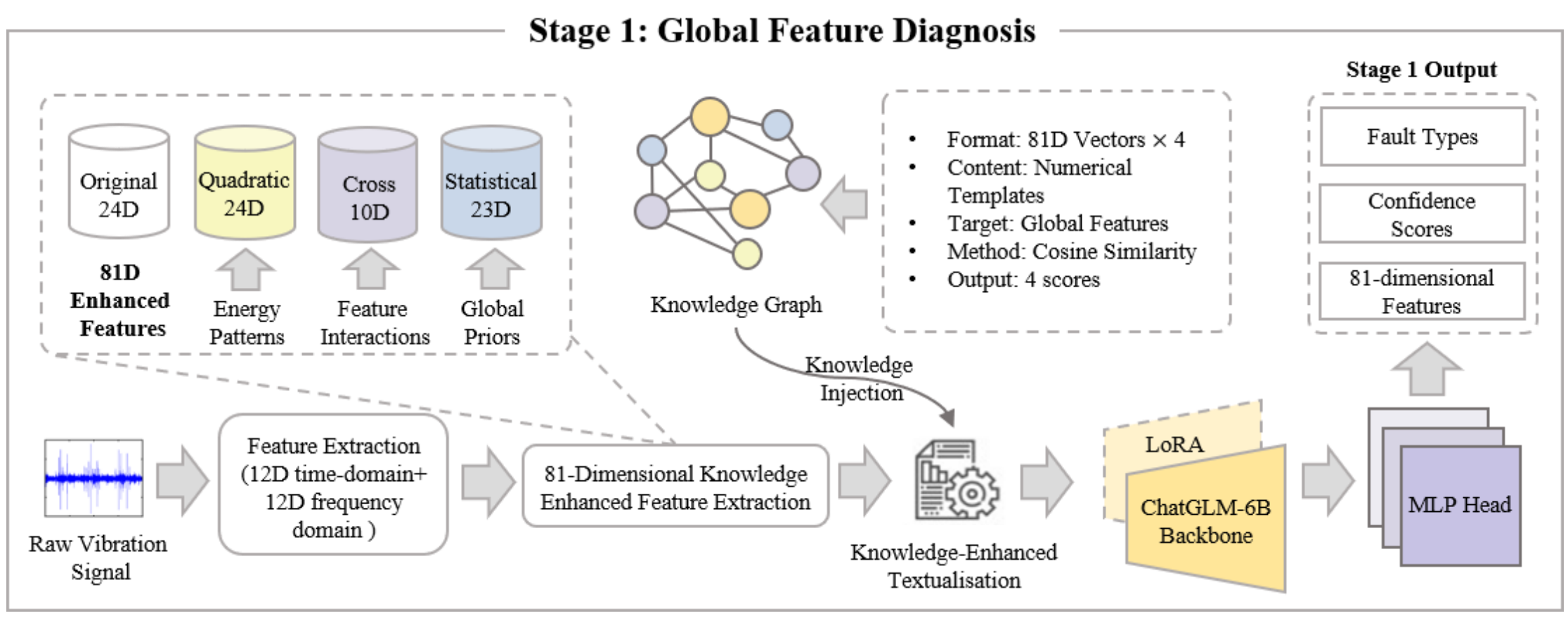

**Fig. 2.** Knowledge-Enhanced Feature-Level Diagnostic Architecture.

Stage 1 employs an 81-dimensional feature vector designed to capture multi-scale fault signatures across complementary signal domains. The feature space is structured as:

$$\boldsymbol{F} = \left[\boldsymbol{F}_{time}, \boldsymbol{F}_{freq}, \boldsymbol{F}_{wavelet}\right] \in \mathbb{R}^{81}, \tag{3}$$

where $\boldsymbol{F}_{time} \in \mathbb{R}^{27}$, $\boldsymbol{F}_{freq} \in \mathbb{R}^{27}$, and $\boldsymbol{F}_{wavelet} \in \mathbb{R}^{27}$ denote time-domain, frequency-domain, and wavelet-domain features, respectively. This expands dimensionality, compared to conventional feature sets, incorporates quadratic energy terms and cross-domain interaction components to enhance discriminative capability.

Time-domain features quantify vibration amplitude characteristics and transient events, including root mean square for energy quantification, kurtosis and skewness for impulsive fault detection, and envelope-based metrics for characteristic fault frequency identification. Quadratic energy terms capture damage progression patterns, while cross-interaction features reveal non-linear dependencies indicative of compound fault scenarios.

Frequency-domain features expose periodic modulation patterns characteristic of localized bearing defects. Key components include spectral kurtosis for transient detection, frequency center of gravity for dominant band tracking, and energy concentrations at characteristic fault frequencies including Ball Pass Frequency Inner race (BPFI), Ball Pass Frequency Outer race (BPFO), and Ball Spin Frequency (BSF), establishing direct links to bearing kinematics.

Wavelet-domain features employ multi-level decompositions to isolate transient impulses across time-frequency scales. For each decomposition level, energy, entropy, and variance descriptors characterize the temporal-spectral distribution of fault signatures. This multi-resolution representation proves particularly effective for incipient fault detection, where damage manifests as subtle coefficient perturbations.

To enhance interpretability, this research integrates KG priors encoding prototypical fault patterns. For each feature vector *F*, similarity scores to fault prototypes are computed as:

$$s_k = \frac{\langle \boldsymbol{F}, \boldsymbol{w}_k \rangle}{(\|\boldsymbol{F}\| \|\boldsymbol{w}_k\|)}, \tag{4}$$

where $\boldsymbol{w}_k \in \mathbb{R}^{81}$ represents the prototype weight vector for fault class *k*, and $s_k \in [-1,1]$ quantifies the alignment between observed features and expected patterns through their inner product $\langle F, w_k \rangle$. These scores serve as soft constraints, guiding subsequent textual interpretation towards physically plausible fault hypotheses.

### 3.2.1 Textualization and LLM-Based Diagnosis

To leverage the contextual reasoning capabilities of LLMs, the 81-dimensional feature vector *F* is converted into structured textual descriptions. This representation enables the model to reason about feature interdependencies. For instance, recognizing that elevated kurtosis concurrent with high BPFI energy strongly suggests inner race defects.

The diagnostic model employs LLM (ChatGLM2-6B in Stage 1), adapted through LoRA for parameter-efficient fine-tuning. LoRA inserts trainable low-rank matrices $\Delta W = \boldsymbol{BA}$ (where $\boldsymbol{B} \in \mathbb{R}^{(d \times r)}, \boldsymbol{A} \in \mathbb{R}^{(r \times d)}$ and $r \ll d$ ) into attention projection layers, reducing trainable parameters to less than 0.1% of the total model while preserving pre-trained knowledge.

For classification, a multi-layer perceptron (MLP) head processes the final token representation:

$$\boldsymbol{h}_{final} \rightarrow MLP(\boldsymbol{h}_{final}) \rightarrow z \in \mathbb{R}^{C}, \tag{5}$$

where $\boldsymbol{h}_{final}$ denotes the final token's encoding after ChatGLM processing of the textualized feature description. This single vector captures global semantic understanding of the statistical features and $C$ represents the number of fault classes, and $z$ contains the class logits. The model is optimized using cross-entropy loss with class-balanced weights to address sample imbalances, employing AdamW optimizer with cosine annealing scheduling.

Stage 1 outputs comprise: (i) predicted fault class probabilities $\boldsymbol{P}_1(y|\boldsymbol{F}) \in \mathbb{R}^{C}$, (ii) the most likely fault type $\hat{y}_1 = \arg\max \boldsymbol{P}_1(y|\boldsymbol{F})$, which serves as the prior $p_1$ for Stage 2 knowledge-guided attention modulation, and (iii) the feature vector $\boldsymbol{F}$ for subsequent multi-modal fusion.

### 3.3 Stage 2: local patch diagnosis

Stage 2 implements a patch-level diagnostic pathway that complements Stage 1 by refining global predictions through fine-grained local signal analysis. As illustrated in Fig. 3, raw vibration signals are first segmented into overlapping patches via sliding window operations. These patches then undergo knowledge-guided embedding, where value embeddings, position embeddings, and knowledge modulation vectors are dynamically weighted based on attention scores derived from a Fault Knowledge Base. This knowledge base, structured as physical diagnostic rules, is queried using Stage 1's feature-level predictions as priors, generating eight attention weights that modulate the embedding process. The modulated patch embeddings are subsequently processed through a 12-layer LoRA-adapted GPT-2 model, referred to as GPT-2-12layer, and a regularised linear classifier produces refined fault predictions alongside patch-level feature representations. Critically, Stage 2 operates directly on numerical signal patches without textualization, thereby preserving temporal fidelity while incorporating semantic guidance from Stage 1 through attention-based knowledge modulation.

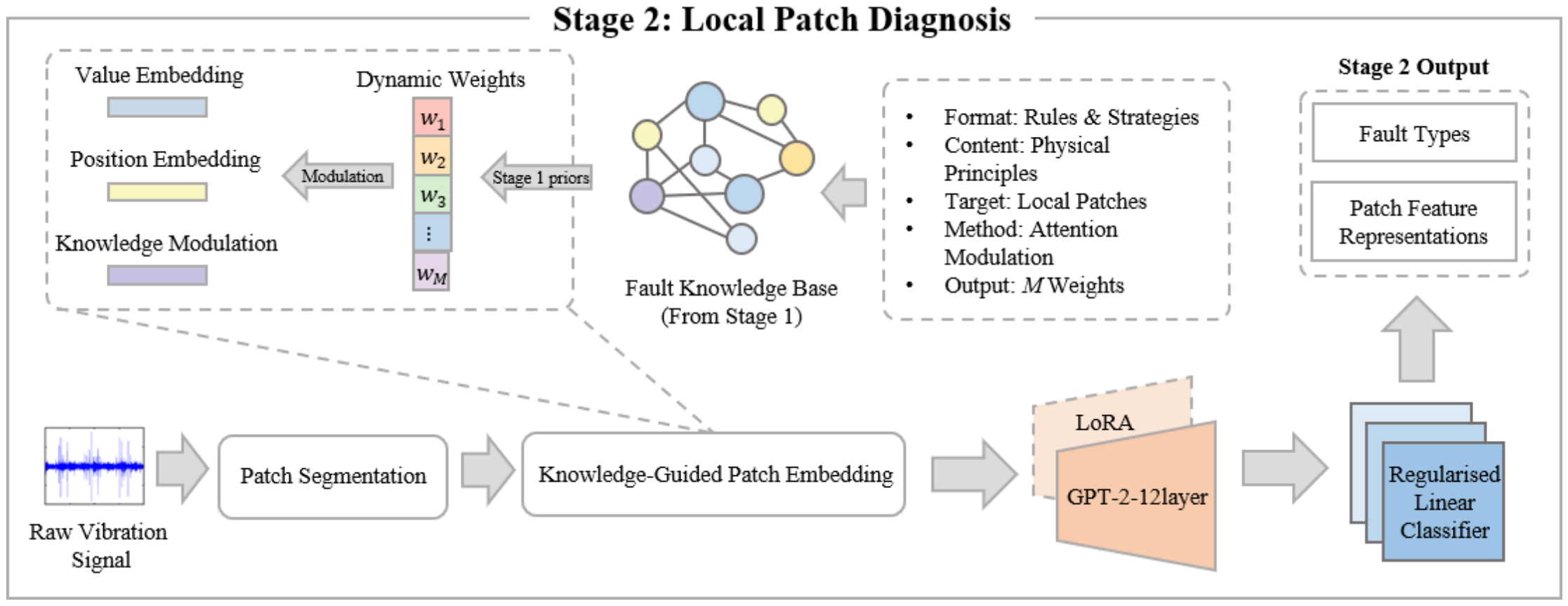

**Fig. 3.** Knowledge-Guided Patch Embedding Diagnosis Architecture.

### 3.3.1 Patch-based signal representation

Global features capture statistical regularities but inherently sacrifice fine-grained temporal structures essential for disambiguating subtle fault signatures. To preserve local waveform morphology, this research adopts a patch-based representation inspired by vision transformers.

The input signal $x(t) \in \mathbb{R}^N$ is segmented into $M$ overlapping patches using a sliding window with patch length $L$ and stride $s$:

$$\boldsymbol{p}_i = \boldsymbol{x}[(i-1)\times s+1:(i-1)\times s+L],\ i=1,\dots,M\ , \tag{6}$$

where $M = \left\lfloor \frac{(N-L)}{s} \right\rfloor + 1$ denotes the total number of patches, and $\boldsymbol{p}_i \in \mathbb{R}^L$ denotes the $i^{th}$ patch. Overlapping patches preserve temporal continuity and capture transient impulses that may span patch boundaries. The patch length $L$ is selected to encompass multiple characteristic fault periods, while the stride $s$ balances computational efficiency and temporal resolution.

To facilitate cross-dataset generalization, per-sample instance normalization is applied before segmentation:

$$\boldsymbol{x}_{norm} = \frac{(\boldsymbol{x}-\mu_x)}{\sqrt{(\sigma_x^2+\varepsilon)}}\ , \tag{7}$$

where $\mu_x$ and $\sigma_x^2$ denote sample-specific mean and variance, and $\varepsilon$ prevents numerical instability [29]. This normalization removes dataset-specific amplitude biases while preserving relative signal morphology.

### 3.3.2 Knowledge-guided patch embedding

The core innovation of Stage 2 lies in modulating patch embeddings based on fault mechanism priors derived from Stage 1. Standard approaches apply uniform embeddings:

$$\boldsymbol{E}_i = Linear(\boldsymbol{p}_i + \boldsymbol{PE}_i), \tag{8}$$

where $Linear(\cdot)$ projects patches to hidden dimension $d$ ($d$=768 for GPT-2), and $\boldsymbol{PE}_i \in \mathbb{R}^d$ denotes positional embeddings. However, bearing fault physics dictates that different spatial regions carry unequal diagnostic relevance. Inner race faults generate periodic impulses concentrated in signal mid-sections, outer race faults exhibit load-zone-dependent amplitude modulation, while ball faults introduce spatially random perturbations.

To encode this domain knowledge, we define fault-specific attention weight patterns. For inner race faults, center patches receive elevated weights; for outer race faults, sinusoidal modulation reflects load-zone effects:

$$\boldsymbol{w}_i = \boldsymbol{w}_{base} + A \cdot \sin(\frac{2\pi(i-1)}{M}), \tag{9}$$

where $\boldsymbol{w} \in \mathbb{R}^M$ denotes the attention weight vector derived from Stage 1 prior $p_1$, $\boldsymbol{w}_{base}$ represents the baseline weight, and $A$ is the modulation amplitude. For normal and ball faults, uniform weights reflect spatially invariant characteristics. The modulated embeddings are computed as:

$$\tilde{\boldsymbol{E}}_i = \boldsymbol{E}_i \odot \boldsymbol{w}_i, \tag{10}$$

where $\odot$ denotes element-wise multiplication with broadcasting. This mechanism amplifies fault-relevant patches while incorporating mechanism-specific inductive biases, effectively constraining the hypothesis space and accelerating convergence.

### 3.3.3 Transformer backbone and classification

The modulated embeddings $\{\tilde{\boldsymbol{E}}_i\}_{i=1}^M$ are processed by a 12-layer GPT-2 model fine-tuned via LoRA, which inserts trainable low-rank matrices $\Delta W$ into attention layers, reducing trainable parameters to less than 1% while preserving pre-trained knowledge. The causal self-attention mechanism captures temporal dependencies across patches.

Final hidden states $\{\boldsymbol{h}_i\}_{i=1}^M$ are aggregated via mean pooling:

$$\boldsymbol{h}_{pooled} = (\frac{1}{M})\sum_{i=1}^{M} \boldsymbol{h}_i \in \mathbb{R}^d . \tag{11}$$

A classification head maps the pooled representation to fault probabilities $\boldsymbol{P}_2(y|\{\boldsymbol{p}_i\}, p_1)$. Layer normalization within the model incorporates learnable affine parameters $\boldsymbol{\gamma}, \boldsymbol{\beta} \in \mathbb{R}^d$ to adapt activation distributions across datasets:

$$y = LayerNorm(\boldsymbol{x}) \odot \boldsymbol{\gamma} + \boldsymbol{\beta}, \tag{12}$$

Training employs cross-entropy loss with class-balanced weighting, optimized using AdamW with cosine annealing scheduling. Stage 2 outputs comprise the fault class probabilities $\boldsymbol{P}_2(y|\{\boldsymbol{p}_i\}, p_1)$ and patch embeddings $\{\tilde{\boldsymbol{E}}_i\}_{i=1}^{M}$ for subsequent fusion.

## 3.4 Stage 3: multi-modal fusion

The Fusion stage unifies multi-scale diagnostic information through heterogeneous token integration. As illustrated in Fig. 4, three modality tokens are constructed: Feature Tokens encoding Stage 1's 81-dimensional global features, Patch Tokens representing Stage 2's local temporal embeddings, and a KG Token capturing domain expertise. Critically, the KG Token is explicitly concatenated during training to guide knowledge integration, but becomes implicit at inference as the learned knowledge resides within model parameters, removing external dependency. The concatenated token sequence is processed by a 6-layer LoRA-adapted GPT-2 model named as GPT-2-6layer, followed by global average pooling and an 11-layer MLP head for fault classification. This design achieves complementary fusion of global patterns, local transients, and physical priors while ensuring deployment efficiency.

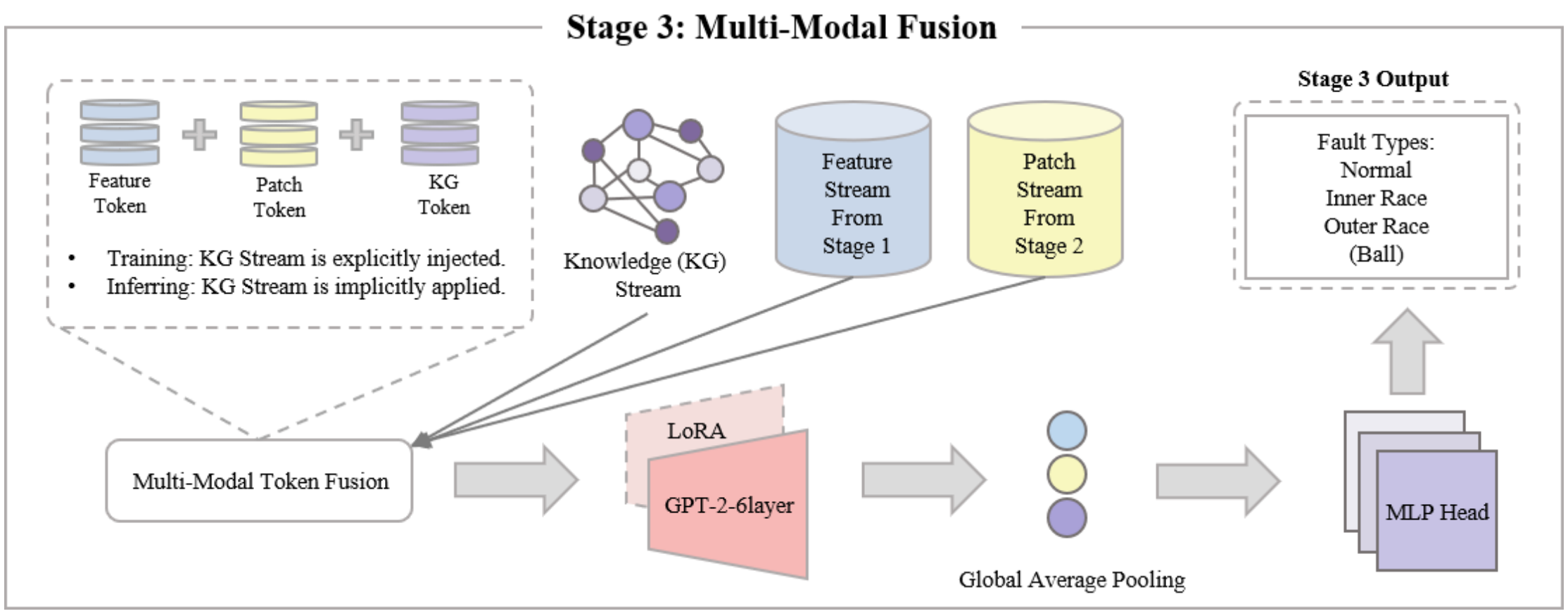

**Fig. 4.** Multi-modal Token Fusion Diagnosis Architecture.

Stages 1 and 2 provide complementary diagnostic perspectives: Stage 1 offers computational efficiency and global pattern recognition, while Stage 2 delivers enhanced accuracy through local detail modelling. However, independent deployment requires sequential inference, complicating real-time deployment and preventing joint optimization of feature and patch representations.

The Fusion stage addresses these limitations by constructing a unified multi-modal architecture that processes features, patches, and external knowledge within a single forward pass, enabling end-to-end gradient flow and learned feature-patch complementarity.

Three parallel encoders project heterogeneous inputs into a shared token space:

**Feature Stream:** The 81-dimensional feature vector $\boldsymbol{F}$ undergoes layer normalization and linear projection:

$$t_{\{global\}} = Linear(LayerNorm(\boldsymbol{F})) \in \mathbb{R}^d, \tag{13}$$

**Patch Stream:** Each patch $\boldsymbol{p}_i$ is embedded with positional encoding:

$$t_{\{patch,i\}} = Linear(\boldsymbol{p}_i) + \boldsymbol{PE}_i \in \mathbb{R}^d,\ i = 1,\ldots,M, \tag{14}$$

**Knowledge Stream:** KG similarity scores $\boldsymbol{s}$ are projected to the token space:

$$t_{\{KG\}} = Linear(\boldsymbol{s}) \in \mathbb{R}^d, \tag{15}$$

where $\boldsymbol{s}$ encodes cosine similarities to fault prototypes (c.f. (4)). These tokens are concatenated to form a multi-modal sequence:

$$\boldsymbol{T} = [t_{\{global\}}, t_{\{patch,1\}}, \ldots, t_{\{patch,M\}}, t_{\{KG\}}] \in \mathbb{R}^{(M+2)\times d}, \tag{16}$$

comprising ($M+2$) tokens with $d$-dimensional embeddings.

A lightweight 6-layer GPT-2 model with LoRA adaptation processes the multi-modal sequence:

$$\boldsymbol{H} = Transformer(\boldsymbol{T}) \in \mathbb{R}^{(M+2)\times d}, \tag{17}$$

where $\boldsymbol{H}$ denotes a matrix containing hidden states from all tokens extracted from the shallower 6-layer GPT-2 backbone. Mean pooling across tokens yields a global representation:

$$\boldsymbol{h}_{\{pooled\}} = \frac{1}{M+2}\sum_{j=1}^{(M+2)} \boldsymbol{h}_j \in \mathbb{R}^d, \tag{18}$$

where $\boldsymbol{h}_j$ represents the $j^{th}$ token's hidden state. A classification head maps $\boldsymbol{h}_{\{pooled\}}$ to fault probabilities $\boldsymbol{P}_{\{final\}}(y|\boldsymbol{F}, \{\boldsymbol{p}_i\}, \boldsymbol{s})$.

**Knowledge injection strategy**: During training, $t_{\{KG\}}$ is explicitly included, providing direct supervision for knowledge integration. During inference, the model operates without $t_{\{KG\}}$, as knowledge becomes internalized in learned weights. This design enables knowledge injection during training while eliminating external dependencies during deployment.

**Training objective:** The model is optimized using cross-entropy loss with class-balanced weighting to address data imbalance:

$$\mathcal{L} = -\frac{1}{N}\sum_{i=1}^{N}\sum_{c=1}^{C} w_c \cdot y_{\{ic\}} \cdot \log(p_{\{ic\}}), \tag{19}$$

where $w_c$ denotes class-specific weights, $y_{\{ic\}} \in \{0,1\}$ represents ground truth labels, N denotes the batch size, $C$ represents the number of fault classes, and $p_{\{ic\}} \in \{0,1\}$ denotes predicted probabilities. AdamW optimizer with cosine annealing scheduling and early stopping is employed to prevent overfitting.

**Progressive architecture advantages:** The sequential design provides interpretable intermediate outputs, where Stage 1 prior $p_1$ guides Stage 2 attention modulation, enabling human-in-the-loop verification critical for industrial deployment. The unified Fusion stage integrates global features ( $\boldsymbol{F}$ ), local patches ( $\{\boldsymbol{p}_i\}$ ), and domain knowledge ( $\boldsymbol{s}$ ) by processing their token representations through a shared Transformer backbone, achieving both high accuracy and interpretability while supporting flexible deployment strategies ranging from rapid Stage 1 screening to comprehensive Fusion analysis.

## 4. Experimental Studies

### 4.1 Experimental setup

#### 4.1.1 Dataset selection and characteristics

To comprehensively evaluate the proposed framework's diagnostic performance and generalization capability, four publicly available bearing datasets with complementary characteristics were selected, as summarized in Table 1.

The CWRU dataset [31] serves as a controlled benchmark with balanced four-class samples acquired at 12 kHz under stable laboratory conditions. The primary diagnostic challenge lies in distinguishing ball faults from inner race faults due to similar frequency characteristics. The JNU dataset [32] introduces rotational speed variability, with samples collected at 600, 800, and 1000 rpm, thereby testing cross-speed generalization under frequency shifts. The PU dataset [33] represents real-world industrial scenarios with heterogeneous operating conditions including multiple bearing types, varying radial loads, and diverse rotational speeds. The MFPT dataset [34] presents dual challenges: severe class imbalance with limited inner race fault samples and mixed sampling rates.

Table 1. Characteristics of bearing datasets

| Dataset | Classes | Sampling Rate (kHz) | Primary Diagnostic Challenge |
| --- | --- | --- | --- |
| CWRU | 4 | 12 | Ball-Inner Race similarity |
| JNU | 4 | 50 | Cross-speed generalization |
| PU | 3 | 64 | Heterogeneous operating conditions |
| MFPT | 3 | 48.8/97.6 | Class imbalance, mixed sampling rates |

#### 4.1.2 Implementation details

The three-stage framework was implemented using LoRA-based parameter-efficient fine-tuning to adapt pre-trained LLMs while preserving generalization capability. Table 2 details the configuration for each stage.

Stage 1 employs ChatGLM2-6B to process 81-dimensional knowledge-enhanced features extracted from time, frequency, and wavelet domains. Stage 2 utilizes GPT-2-12layer to encode raw signals segmented into eight overlapping patches of 256 sampling points each. The Fusion stage integrates multi-modal tokens through GPT-2-6layer, unifying feature-level representations, patch-level embeddings, and KG information.

All stages adopt LoRA rank $r$=8, with scaling factors α and learning rates optimized for each module's convergence characteristics. Training was conducted using the AdamW optimizer with batch size 32. Stage 1 and Stage 2 were trained for 30 epochs, while the Fusion stage required 50 epochs to ensure convergence in the unified representation space.

Table 2. Model configuration for three-stage framework

| **Parameter** | **Stage 1** | **Stage 2** | **Fusion** |
|---|---|---|---|
| Backbone | ChatGLM2-6B | GPT-2-12layer | GPT-2-6layer |
| Parameters | 6.2 billion | 124 million | 62 million |
| Input Modality | 81-dimension features | 8×256 patches | Multi-model tokens |
| LoRA rank ($r$) | 8 | 8 | 8 |
| LoRA scaling ($\alpha$) | 32 | 32 | 16 |
| Learning rate | $1\times10^{-5}$ | $2\times10^{-4}$ | $5\times10^{-4}$ |
| Training epochs | 30 | 30 | 50 |

Model performance was assessed using overall classification accuracy, per-class precision, recall, and F1-score. Confusion matrices were employed to visualize classification patterns and identify systematic misclassification trends. All experiments were conducted using 5-fold cross-validation, with mean and standard deviation reported to quantify performance stability.

### 4.2 Overall performance and progressive improvement

Table 3 presents classification accuracy across the three-stage framework for all four datasets. The progressive architecture demonstrates consistent performance gains, with overall average accuracy improving from 94.33% at Stage 1 to 97.90% at Stage 2 and 98.49% at Fusion. Stage 2 outperforms Stage 1 by an average of 3.57 percentage points, validating the effectiveness of knowledge-guided patch embedding defined in (6)-(9) of Section 3. Fusion provides additional gains on three of four datasets, ranging from 0.30% to 2.01%.

Table 3. Classification accuracy across three-stage framework

| **Dataset** | **Stage 1** | **Stage 2** | **Fusion** | **Stage 1→2** | **Stage 2→F** |
|---|---|---|---|---|---|

| CWRU | 95.59% | 97.61% | 98.19% | +2.02 | +0.58 |
|---|---|---|---|---|---|
| JNU | 89.69% | 99.77% | 99.27% | +10.08 | -0.50 |
| PU | 92.98% | 95.05% | 97.06% | +2.07 | +2.01 |
| MFPT | 99.07% | 99.15% | 99.45% | +0.08 | +0.30 |
| Average | 94.33% | 97.90% | 98.49% | +3.57 | +0.59 |

Note: Stage1→2 and Stage 2→F columns indicate percentage point changes between consecutive stages.

Stage 1 establishes a 94.33% baseline, with lower performance on JNU dataset at 89.69% indicating limited rotational speed invariance of handcrafted features. Stage 2 yields substantial improvements, particularly on JNU with a 10.08% gain where attention modulation focuses on fault-indicative impulses despite frequency-domain shifts. The modest MFPT improvement of 0.08 percentage points suggests Stage 1 features already capture sufficient information under mixed sampling rates. Fusion achieves the largest gain of 2.01 percentage points on PU, where complementary feature and patch representations compensate for operational heterogeneity. On JNU, Fusion accuracy decreases slightly to 99.27%, indicating saturation where the 6-layer backbone provides no additional benefit beyond 99.77% in Stage 2.

Dataset-specific patterns reveal framework adaptability. MFPT and JNU exceed 99% accuracy, with mixed sampling rates encouraging scale-invariant learning and cross-speed training promoting rotational-order-independent signatures. PU exhibits 97.06% accuracy despite extreme operational variability, demonstrating practical industrial applicability. CWRU shows steady improvement from 95.59% to 98.19%, confirming cumulative benefits under stable conditions. Three observations emerge: knowledge-guided attention provides the largest improvement of 3.57 percentage points, particularly benefiting domain shift scenarios; multi-modal fusion contributes most under heterogeneous conditions where complementary representations mitigate individual limitations; and saturation occurs beyond 99.5% single-stage accuracy, indicating the Fusion backbone prioritizes integration over capacity.

### 4.3 Per-class performance analysis

Table 4 presents per-class precision, recall, and F1-scores for the Fusion stage across all four datasets, formatted as Precision/Recall/F1-Score for each fault category. All datasets achieve weighted average F1-scores exceeding 0.970. CWRU demonstrates balanced performance across all classes, with Ball class achieving perfect scores and Inner race showing 0.930/0.999/0.963 despite mechanical similarity to Ball faults. JNU exhibits near-perfect metrics exceeding 0.98 F1-score for all classes, validating cross-speed generalization under 600-1000 rpm variations. PU maintains robust performance of 0.971 average F1-score despite extreme operational heterogeneity, while MFPT achieves 0.994 despite severe class imbalance and mixed sampling rates.

Table 4. Per-class metrics for fusion model

| **Dataset** | **Normal** | **Inner Race** | **Outer Race** | **Ball** | **Weighted Average** |
|---|---|---|---|---|---|
| CWRU | 1.000/1.000/1.000 | 0.930/0.999/0.963 | 0.985/0.957/0.971 | 1.000/0.962/0.980 | 0.983/0.983/0.982 |
| JNU | 0.997/0.999/0.998 | 0.994/0.970/0.982 | 0.989/0.995/0.992 | 0.981/0.995/0.988 | 0.993/0.993/0.993 |
| PU | 0.995/0.973/0.984 | 0.933/0.983/0.957 | 0.967/0.962/0.965 | - | 0.971/0.971/0.971 |
| MFPT | 1.000/0.999/1.000 | 1.000/0.965/0.982 | 0.981/1.000/0.990 | - | 0.995/0.994/0.994 |

Note: The format is Precision/Recall/F1-Score. Weighted average computed across all classes.

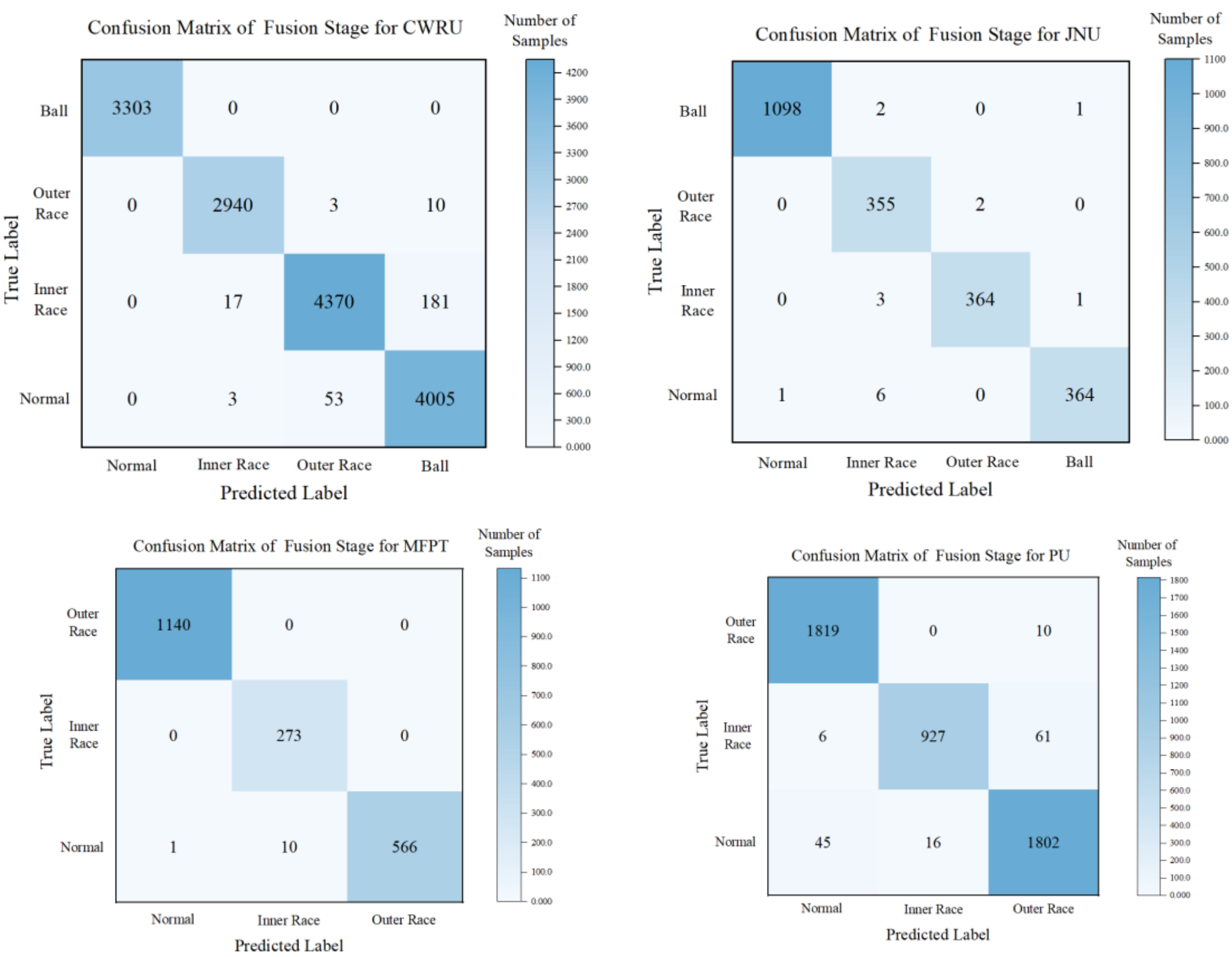

**Fig. 5.** Confusion Matrices for Four Datasets.

Fig. 5 presents confusion matrices revealing dataset-specific diagnostic patterns. CWRU exhibits strong diagonal dominance with primary confusions between Inner and Ball classes, attributable to shared impulsive characteristics under similar mechanical configurations. JNU demonstrates near-perfect classification with minimal off-diagonal elements, confirming robust cross-speed generalization where morphological features outweigh frequency-domain variations. PU shows moderate confusion between Normal and Outer classes under high radial loads, where load-induced vibrations resemble localized defects. MFPT achieves exceptional purity despite 2:1 sampling rate variation and severe class imbalance, with Outer race demonstrating perfect discrimination.

## 4.4 Computational efficiency analysis

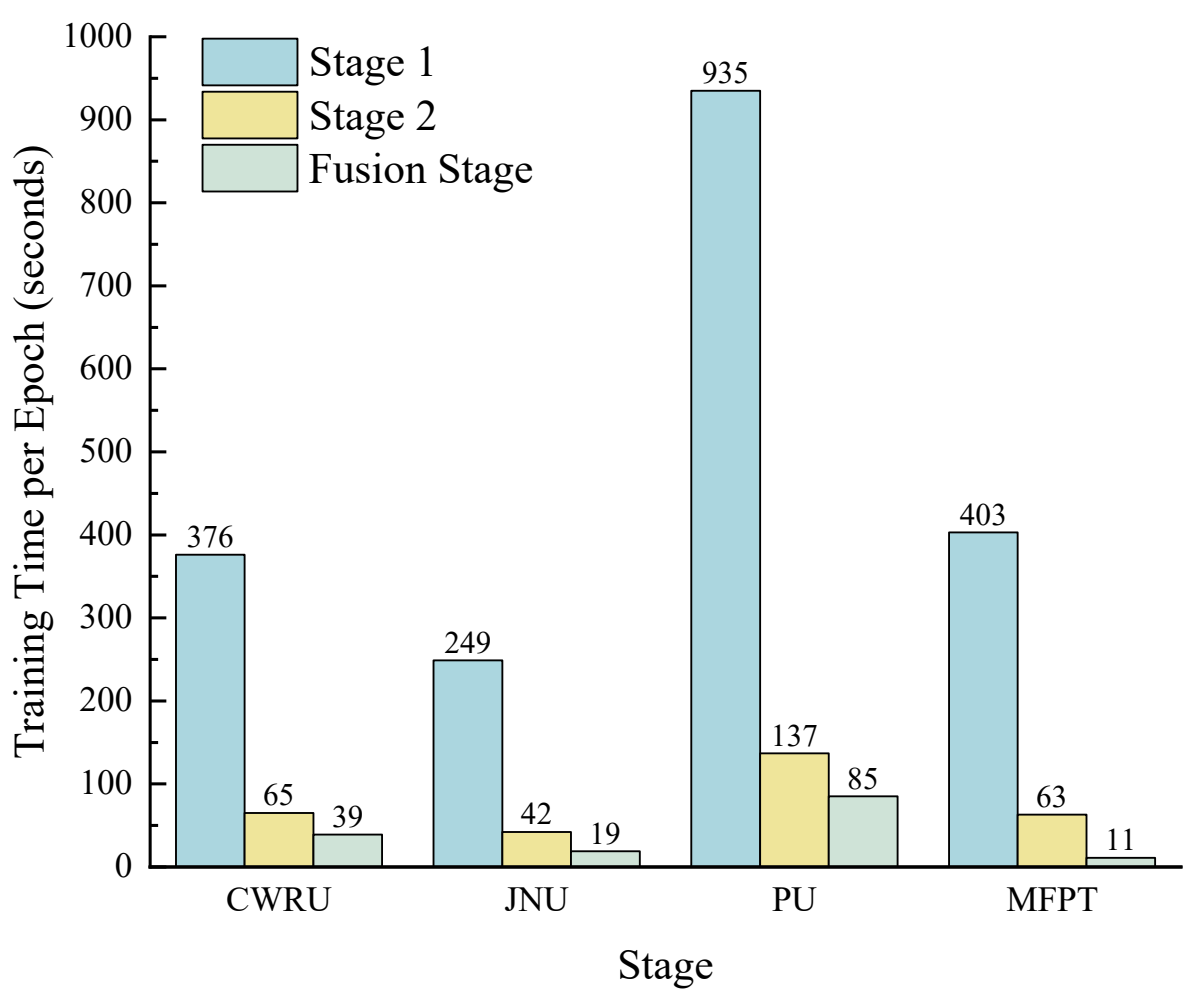

**Fig. 6**. Average Training Time per Epoch of Three Stages.

Fig. 6 presents training time per epoch across the three stages for all four datasets. Stage 1 exhibits the longest training times averaging 491 seconds per epoch, attributed to the ChatGLM-6B backbone with 6.2 billion parameters despite LoRA reducing trainable parameters to 6 million. Stage 2 achieves 6.5-fold speedup with 76-second average using the GPT-2-12layer backbone of 124 million parameters. Fusion stage demonstrates the shortest training time of 39 seconds per epoch on average, representing 12.6-fold speedup over Stage 1 while simultaneously improving accuracy by 4.16 percentage points from 94.33% to 98.49%. This efficiency-performance trade-off validates that diagnostic accuracy does not require billion-parameter models when knowledge-guided mechanisms enable effective integration of multi-modal representations. An unexpected training efficiency pattern emerges. CWRU Stage 1 requires 376 seconds per epoch, substantially shorter than PU's 935 seconds, despite CWRU containing approximately 20,000 segments from 233 raw files compared to PU's 37,470 samples from 150 pre-segmented files. This apparent contradiction arises from differing batch size configurations. CWRU employs a batch size of 32 to accelerate convergence over its larger sample pool, resulting in approximately 500 gradient update steps per epoch. In contrast, PU utilizes a batch size of 8 for more granular optimization, necessitating 3,512 steps per epoch, which is seven times more than CWRU. Consequently, while single-step processing times remain comparable between datasets, the step count difference accounts for the observed 2.5-fold disparity in epoch duration.

Training the Fusion stage converges within 50 epochs, requiring approximately 33 minutes total on average across datasets. The LoRA configuration with rank 8 and scaling $\alpha$=16 limits trainable parameters to 0.6 million, enabling fine-tuning on consumer-grade GPUs. The lightweight architecture facilitates deployment in resource-constrained industrial environments without dedicated high-performance computing infrastructure, supporting rapid adaptation to new bearing types or operating conditions.

### 4.5 Multi-modal Interpretability Analysis

To verify that diagnostic decisions are traceable to known bearing fault mechanisms rather than opaque statistical correlations, two complementary analyses are presented: token-level contribution analysis quantifying the relative importance of each information stream, and cross-modal attention visualization revealing how heterogeneous representations interact during inference.

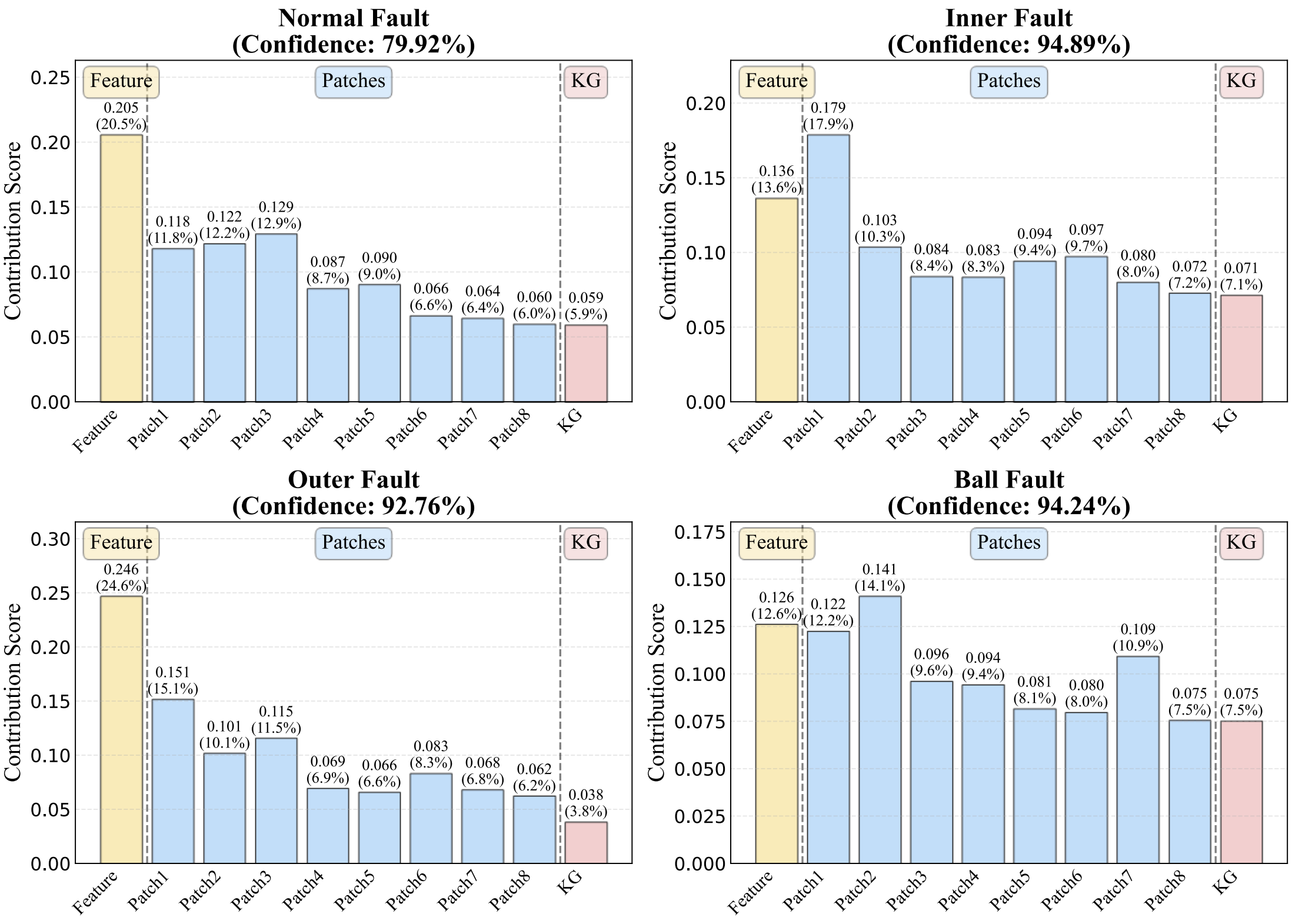

**Fig. 7.** Token Contribution Distribution for Four Fault Categories.

Fig. 7 presents normalized gradient magnitudes for the Feature, Patch, and KG tokens across four fault categories. Patch tokens consistently dominate diagnostic decisions, contributing 71.54% for outer race faults and 79.89% for ball faults, substantially exceeding the 10% uniform baseline. This dominance reflects the impulsive, spatially concentrated nature of bearing fault vibrations, which manifest most distinctively at the patch level. The fault-type dependence of these contributions further aligns with known fault mechanics: ball and inner race faults, which produce higher-frequency periodic impacts due to rolling element–raceway contact geometry, exhibit stronger patch contributions than outer race faults, whose load-zone modulation distributes energy more globally and correspondingly demands higher Feature token reliance of 24.65%. KG contributions range from 3.82% to 7.52%, with impulsive fault types receiving stronger prior support. This adaptive modulation confirms that domain knowledge embedded during training selectively reinforces ambiguous diagnostic signals rather than contributing uniformly, a behavior fundamentally different from static knowledge integration approaches.

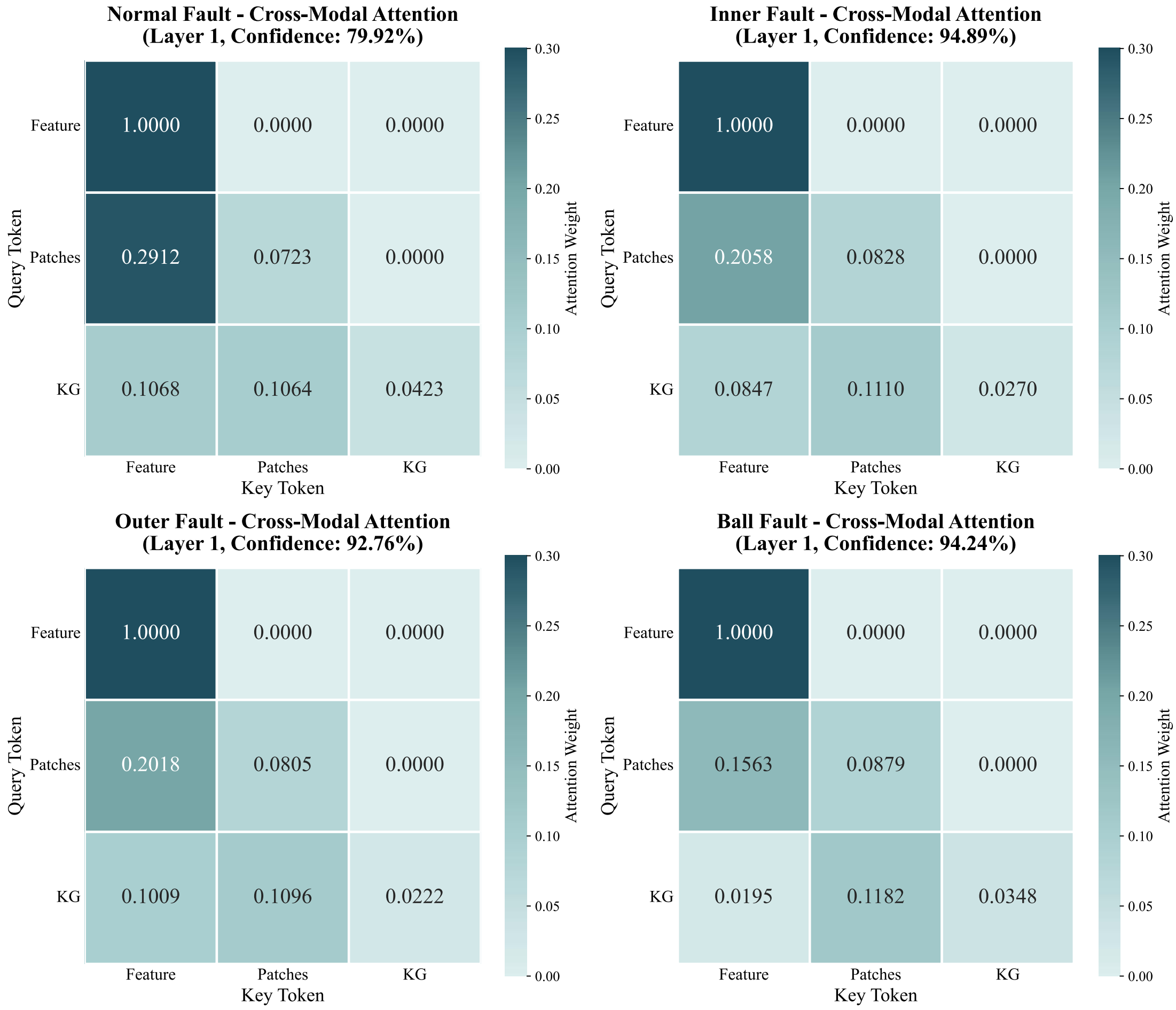

**Fig. 8.** Cross-Modal Attention Matrices.

Fig. 8 presents the cross-modal attention matrices from the first Transformer layer of the Fusion stage, with weights averaged across six heads. Four interaction patterns each admit a physically grounded interpretation. KG-to-Patch attention ranging from 0.1064 to 0.1182 reflects active knowledge validation of local signal morphology, with higher weights for ambiguous patterns confirming that domain knowledge is most intensively recruited when data-driven evidence is less discriminative. KG-to-Feature attention spanning 0.0195 to 0.1068 demonstrates that stationary vibration patterns require greater integration of global statistical descriptors with domain knowledge, consistent with the broader distributional nature of such signatures. Patch-to-Feature attention ranging from 0.1563 to 0.2912 reveals that local signal segments seek global statistical context in proportion to their individual discriminability, while inter-Patch attention from 0.0723 to 0.0879 captures the consistent role of temporal context in characterizing fault periodicity across all fault types.

These two analyses provide mutually reinforcing evidence for physics-consistent multi-modal reasoning: the token contributions establish which information stream dominates each fault decision, while the attention matrices reveal the mechanism by which complementary streams are integrated. The observed patterns confirm that the proposed framework encodes genuine bearing dynamics expertise rather than statistical co-occurrence, providing the diagnostic interpretability necessary for practitioner trust in industrial deployment.

## 5. Conclusion

This paper proposed a three-stage progressive physics-guided vibration signal processing framework addressing three measurement challenges in bearing fault diagnosis: the scale conflict between global feature efficiency and local transient fidelity, insufficient traceability of measurement features to fault physics, and integration insufficiency of multi-source measurement information across diagnostic scales. A physically grounded 81-dimensional measurement descriptor traceable to characteristic defect frequencies enables auditable fault screening. A fault-adaptive signal segmentation mechanism directs analytical attention toward fault-relevant waveform regions guided by Stage 1 priors. Structured fault mechanism knowledge is encoded implicitly in model parameters during training, eliminating external dependencies and enabling autonomous field operation. Validation on four benchmark datasets demonstrated 98.49% average accuracy, a 4.16 percentage point gain over the Stage 1 baseline of 94.33%, with 12.6-fold computational cost reduction relative to signal-level baselines. Stage 1 screening latency of approximately 20 milliseconds per sample confirms suitability for resource-constrained industrial monitoring. Interpretability analysis confirmed that diagnostic feature activations align with established bearing fault mechanics, supporting measurement traceability in safety-critical systems.

Current validation is limited to within-dataset evaluation. Cross-dataset transfer learning constitutes the primary direction for future work, alongside adaptive fusion for compound faults, fault severity monitoring, and lightweight variants for edge deployment.